\documentclass[letterpaper, 10 pt, conference]{ieeeconf}  

\IEEEoverridecommandlockouts                              

\usepackage{amsmath}
\usepackage{amssymb}
\usepackage{booktabs}
\usepackage{tikz}
\usetikzlibrary{arrows.meta,positioning,fit,calc}
\usepackage{cite}
\usepackage{amsmath,amssymb,amsfonts}
\usepackage{graphicx}
\usepackage{textcomp}
\usepackage{xcolor}
\usepackage{subcaption}
\usepackage{float}
\usepackage{balance}
\usepackage[linesnumbered,ruled,vlined]{algorithm2e}
\DontPrintSemicolon
\usepackage{url}
\usepackage{svg}
\usepackage{hyperref}
\usepackage{verbatim}
\usepackage{multirow}
\usepackage{spverbatim}
\usepackage{fancyvrb}
\usepackage{comment}
\usepackage{float}
\usepackage{todonotes}
\def\BibTeX{{\rm B\kern-.05em{\sc i\kern-.025em b}\kern-.08em
    T\kern-.1667em\lower.7ex\hbox{E}\kern-.125emX}}

\title{\LARGE \bf Ontology-Guided Reasoning for Affordance-Based Explanations of Robot Navigation}

\author{Amar Halilovic$^{1}$, Vahidin Hasic$^{2}$ and Senka Krivic$^{2}$
\thanks{*This work was not supported by any organization}
\thanks{$^{1}$Amar Halilovic is with the Institute of Artificial Intelligence,
        Ulm University, 89081 Ulm, Germany,
        {\tt\small amar.halilovic@uni-ulm.de}}%
\thanks{$^{2}$Vahidin Hasic and Senka Krivic are with the Faculty of Electrical Engineering, University of Sarajevo,
        71000 Sarajevo, BiH,
        {\tt\small {\{vahidin.hasic,senka.krivic\}}@etf.unsa.ba}}%
}

\begin{document}

\maketitle

\begin{abstract}
This paper proposes ontology-guided reasoning for affordance-based explanations of robot navigation. 
In human environments, it is not sufficient for a robot to detect that its route is blocked. It must also reason about what nearby objects afford, which state changes are possible, and which of these changes would allow it to continue safely. 
We address this problem by representing nearby entities, their affordances, affordance states, and qualitative spatial relations in a local affordance ontology and by evaluating hypothetical object--affordance state changes as candidate explanation factors. 
This yields explanations that are not only semantically grounded but also actionable. 
We instantiate the approach in a lightweight benchmark centered on a robot librarian scenario and evaluate it on procedurally generated navigation cases. 
The results show that ontology-guided reasoning identifies relevant explanation factors more accurately than a semantic-only baseline and remains robust as semantic clutter increases. 
Overall, the paper argues that affordance ontologies can serve not merely as semantic descriptions of the environment, but as reasoning foundations for explainability and reliable robot autonomy.
\end{abstract}




\section{Introduction}
Affordances, initially introduced in ecological psychology by Gibson~\cite{gibson1977affordances}, are the properties of objects that enable or constrain possible actions. 
In robotics, this idea is highly relevant, since robots need to understand not only what is present in the environment, but also what can be done with it. 
A door can be opened, a chair moved, and a cart pushed away. In other words, the environment does not merely contain objects, but it also contains actionable possibilities.

Path planning is at the heart of autonomous robotic systems: it involves formulating a robot's movement through an environment to fulfill a task. 
However, path planners often behave like black boxes, yielding a path or a failure message without clarifying why a particular route could not be followed. 
This lack of transparency becomes especially problematic in indoor social settings, where nearby humans may be able to help the robot, but only if the robot can communicate what is blocking it and what change in the environment would resolve the problem. 
Explanations are therefore a promising way to make robot navigation more understandable and transparent, and prior work has shown that explanations can positively affect trust and understanding in human--robot interaction (HRI)~\cite{wachter2017robotics,stange2020self}.

\begin{figure}[t]
\centering
\includegraphics[width=0.25\textwidth]{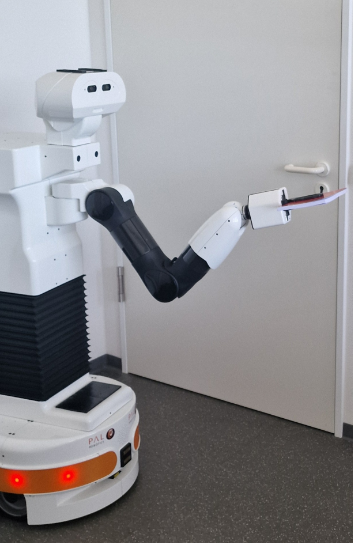}
\caption{A robot librarian holding a book navigates a library environment, where nearby objects may obstruct the path and prompt an explanation.}
\label{fig:mot}
\end{figure}

We approach this problem in the context of a robot librarian. 
Consider a scenario in which a robot has to fetch a book for a visitor in a library. 
While navigating toward the corresponding bookshelf, the direct path may become blocked by a chair, a closed cabinet door, a trolley, or even a person temporarily standing in the way. 
In such a case, a useful explanation should not merely state that the path is blocked. 
Instead, it should indicate which nearby object is responsible, what affordance is relevant, and what change would allow the robot to proceed. 
For example, it is more useful to say that a cabinet in front of the robot needs to be opened or that a chair needs to be moved than to merely report that the plan failed. 
Figure~\ref{fig:mot} shows the motivating library scenario: a robot librarian holding a book must continue its task despite nearby objects that may obstruct the route and require explanation.

Although existing feature-based explainability methods can identify which objects or regions of space are important for a navigation outcome, they do not naturally capture which \emph{actionable change} in the environment would help the robot proceed. 
We argue that affordance-based explanations require more than geometric or feature-level attribution: they require an explicit semantic reasoning layer that represents objects, their affordances, and the consequences of changing these states for navigation.
This makes the resulting explanations not only interpretable but also actionable, grounding them not only in post-hoc description but also in an explicit semantic reasoning process.
Figure~\ref{fig:framework} summarizes the resulting reasoning pipeline, from local semantic scene representation to the selection of the top-ranked explanation factors.

The main contributions of this paper are:
\begin{itemize}
    \item We formalize a local affordance ontology and introduce ontology-guided reasoning for affordance-based explanations of robot navigation.
    \item We present a lightweight benchmark~\footnote{All code and evaluation materials used in this work are publicly available at the accompanying \href{https://github.com/ah96/affordance_ontology_guided_reasoning_explanations}{GitHub repository}.} for evaluating ontology-guided affordance reasoning.
    \item We experimentally show that ontology-guided reasoning identifies relevant explanation factors more accurately than a semantic-only baseline and remains robust as semantic clutter increases.
\end{itemize}

\section{Related Work}
Affordances have been extensively used in robotics to support action understanding and interaction in complex environments. 
One of the early works that formalized affordances for robot control is by \c{S}ahin et al.~\cite{sahin2007afford}, while Montesano et al.~\cite{montesano2008learning} developed a probabilistic framework for learning object affordances from action effects. 
Lindner and Eschenbach further developed affordance-based conceptualizations of robot behavior in shared environments and affordance spaces~\cite{lindner2013placement,lindner2017conceptual}. 
In contrast to work focusing primarily on robot interaction capabilities, we are interested in how affordances can be leveraged to explain robot navigation in a semantically meaningful and actionable way.

In explainable robotics, many existing approaches investigate planning failure, contrastive path explanations, or self-explanations of robot behavior~\cite{kwon2018incapability,brandao2021optimality,krarup2021contrastive,stange2020self}. 
Environment-based explanations of motion planning failure have also recently been explored~\cite{liu2024environment}. 
However, much of this work either focuses on geometric changes in the environment or uses semantic information primarily at the presentation level. 
Rather than primarily studying the explanation presentation, we focus on semantic reasoning over an affordance ontology to explain and improve route reliability.

\section{Ontology-Guided Affordance Reasoning}
In the context of this paper, an \emph{ontology} is a structured knowledge representation of the robot's local environment that specifies which objects are present, which properties and affordances they have, what states these affordances are currently in, and how the objects are qualitatively related to the robot. 
The ontology, therefore, serves as the semantic layer on top of the geometric navigation map. 
In our setting, \emph{affordances} are important because they link nearby objects to possible interventions that may restore or improve the robot's route, thereby serving as explanatory factors.

In the library scenario, nearby objects may offer different possibilities: a chair can be moved, a door can be opened, and a person may clear the path by stepping aside. 
To represent such information explicitly, we model the robot's local surroundings as an affordance ontology
\begin{equation}
O_t=\langle X_t, A, S, R \rangle,
\end{equation}
where $X_t$ is the set of object instances currently represented in the robot's local map at time $t$, $A$ is the set of affordance types, $S$ is the set of affordance states, and $R$ is the set of qualitative spatial relations that may hold between objects and the robot. 
The subscript $t$ indicates that the local ontology is updated as the robot moves and as the surrounding scene changes.

We associate each object instance $x_i \in X_t$ with an ontology record
\begin{equation}
\phi_t(x_i) =
\left\langle
\tau(x_i),\,
\mathrm{Aff}(x_i),\,
\sigma_t(x_i),\,
\rho_t(x_i, q_t)
\right\rangle ,
\end{equation}
where $\tau(x_i)$ denotes the semantic object type, $\mathrm{Aff}(x_i) \subseteq A$ denotes the set of affordances associated with the object, $\sigma_t(x_i): \mathrm{Aff}(x_i) \rightarrow S$ denotes the current affordance-state assignment, and $\rho_t(x_i; q_t) \in R$ denotes the qualitative spatial relation between the object and the robot, computed with respect to the robot pose contained in $q_t$. 
We focus on binary affordance states for simplicity, e.g., open/closed or moved/not moved, following the affordance-based explanation setting introduced in our prior work~\cite{halilovic2025affordance}.


The ontology is intentionally local and robot-centered. 
It does not aim to model the entire library at once, but rather the part of the scene currently relevant to navigation and explanation. 
This choice follows the intuition that navigation explanations are usually grounded in a local neighborhood around the robot. 
For instance, when the robot's route fails because of an object directly in front of it, distant objects elsewhere in the library are not explanation-relevant, even if they are part of the global semantic map. 
Locality, therefore, helps both computational efficiency and interpretability.

The ontology also constrains the explanation space. 
Rather than allowing arbitrary explanations about nearby obstacles or arbitrary changes in the environment, it restricts candidate explanations to semantically valid object--affordance--state combinations that are actually supported by the local scene representation. 
In this way, the robot does not reason about unrestricted environment modifications, but rather about admissible interventions.
This constraint is important because it makes the resulting explanations more focused, more actionable, and more consistent with the robot's semantic understanding of the environment.

Given a start state $s$, a goal state $g$, and an ontology-induced navigation map $M(O_t)$, the robot first computes the baseline path cost
\begin{equation}
C_0 = \mathrm{Cost}\bigl(\mathrm{Plan}(s,g,M(O_t))\bigr),
\end{equation}
where $\mathrm{Plan}$ denotes a shortest-path planner and $\mathrm{Cost}$ returns the cost of the resulting path. 
If no feasible path exists, the cost is set to $\infty$.

For every object $x_i \in X_t$ and every currently unfulfilled affordance $a \in \mathrm{Aff}(x_i)$, the reasoner considers a hypothetical affordance-state change
\begin{equation}
h_{x_i,a}: \sigma_t(x_i)(a)=0 \rightarrow 1,
\end{equation}
and recomputes the path cost under the updated ontology:
\begin{equation}
C_{x_i,a} =
\mathrm{Cost}\bigl(\mathrm{Plan}(s,g,M(O_t \oplus h_{x_i,a}))\bigr).
\end{equation}
Here, $\oplus$ denotes an ontology update under a hypothetical intervention, and $M(\cdot)$ denotes the navigation map induced by the corresponding ontology.

We score each candidate object--affordance pair $(x_i,a)$ by
\begin{equation}
\small
U(x_i,a)=
\begin{cases}
\Lambda, & C_0=\infty,\ C_{x_i,a}<\infty,\\
\max(0, C_0-C_{x_i,a}), & C_0,C_{x_i,a}<\infty,\\
0, & \text{otherwise.}
\end{cases}
\label{utility::eq}
\end{equation}
Here, $\Lambda$ is a fixed bonus for restoring feasibility. 
The score, therefore, reflects how helpful a hypothesized affordance-state change is for navigation. 
We rank all candidates by this score and select the highest-ranked object--affordance pairs as explanation factors.

\paragraph*{Library Example}
Assume that the robot is navigating toward a bookshelf to fetch a requested book. 
A chair and a cabinet are located directly in front of the robot, while a cart stands farther away in a side corridor. 
The chair has the affordance \texttt{movable}, the cabinet has affordances \texttt{movable} and \texttt{openable}, and the cart has the affordance \texttt{movable}. 
Let the current affordance states indicate that the chair has not been moved, the cabinet is closed, and the cart has not been moved. 
Suppose that the baseline planner yields $C_0=\infty$, i.e., there is no feasible route to the goal under the current state assignments. 
The reasoner then evaluates hypothetical changes. 
If opening the cabinet yields a feasible path while moving the cart does not change the route, then the pair \emph{cabinet--openable} receives the highest utility. 
This allows the robot to generate an explanation that is both semantically grounded and actionable: \emph{``Please open the cabinet in front of me so I can continue to the bookshelf.''}
This example highlights an important point: two nearby objects need not be equally explanation-relevant. 
What matters is not proximity alone, but whether the corresponding affordance-state change improves or restores the robot's route. 

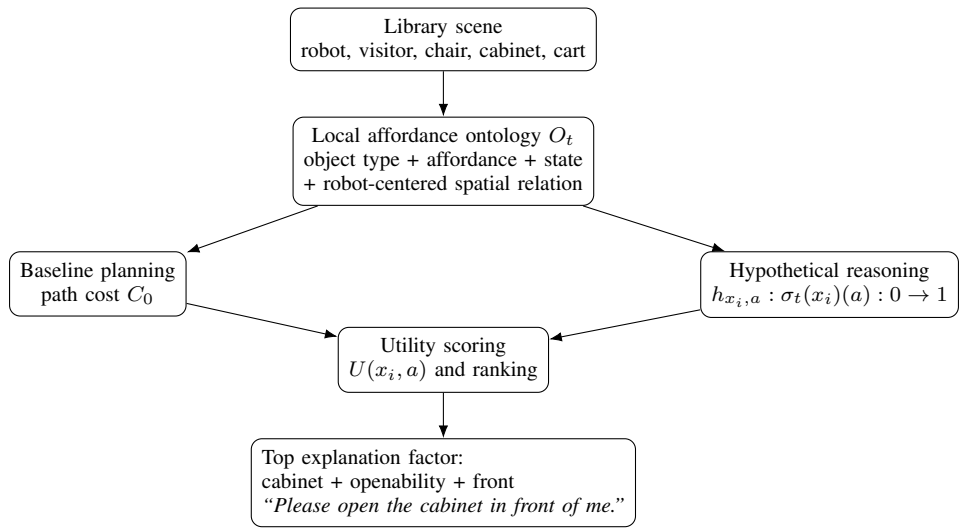
\begin{figure*}[htb]
\centering
\begin{tikzpicture}[>=Latex, node distance=6mm and 6mm, font=\footnotesize,
box/.style={draw, rounded corners, align=center, inner sep=4pt},
smallbox/.style={draw, rounded corners, align=left, inner sep=4pt}]

\node[box] (scene) {Library scene\\robot, visitor, chair, cabinet, cart};

\node[box, below=of scene] (ontology) 
{Local affordance ontology $O_t$\\
object type + affordance + state\\
+ robot-centered spatial relation};

\node[box, below left=of ontology, xshift=-8mm] (planner) 
{Baseline planning\\path cost $C_0$};

\node[box, below right=of ontology, xshift=8mm] (counter) 
{Hypothetical reasoning\\
$h_{x_i,a}:\sigma_t(x_i)(a):0\rightarrow1$};

\node[box, below=16mm of ontology] (score) 
{Utility scoring\\$U(x_i,a)$ and ranking};

\node[smallbox, below=of score] (exp) 
{Top explanation factor:\\
cabinet + openability + front\\
\emph{``Please open the cabinet in front of me.''}};

\draw[->] (scene) -- (ontology);
\draw[->] (ontology) -- (planner);
\draw[->] (ontology) -- (counter);
\draw[->] (planner) -- (score);
\draw[->] (counter) -- (score);
\draw[->] (score) -- (exp);

\end{tikzpicture}
\caption{Ontology-guided affordance reasoning in the library scenario. 
The robot reasons about nearby objects, their affordances, current affordance states, and robot-centered qualitative spatial relations to identify relevant and actionable explanation factors.}
\label{fig:framework}
\end{figure*}

\section{Experimental Design}
We instantiate the proposed method in a lightweight Python benchmark centered on the robot librarian scenario. 
The benchmark mirrors the reasoning process illustrated in Fig.~\ref{fig:framework}: the robot first evaluates the current route, then reasons over hypothetical affordance-state changes, and finally ranks the resulting explanation factors.

Each environment is a grid world with a start location, a goal location representing the requested bookshelf, a direct corridor, and a longer detour corridor. 
Semantic objects \texttt{chair}, \texttt{door}, \texttt{cabinet}, \texttt{person}, and \texttt{cart} are placed in the environment. 
Each object type determines the affordance types considered by the reasoner:
\texttt{chair}   $\mapsto \{$\texttt{movable}$\}$,
\texttt{door}    $\mapsto \{$\texttt{openable}$\}$,
\texttt{cabinet} $\mapsto \{$\texttt{movable},\texttt{openable}$\}$,
\texttt{person}  $\mapsto \{$\texttt{can\_step\_aside}$\}$,
\texttt{cart}    $\mapsto \{$\texttt{movable}$\}$.
To evaluate whether ontology-guided reasoning identifies meaningful explanation factors, we distinguish between \emph{shortcut} objects and \emph{distractor} objects. Shortcut objects are placed on the direct route and constitute the relevant causal factors for route recovery or improvement. Distractor objects are semantically plausible but less relevant to the robot's current route. This design allows us to test whether the method can distinguish relevant explanation candidates from nearby but less important objects.

The planner operates on a semantically induced cost map. Satisfied affordance states are assigned a cost of $1$. A closed door is assigned a cost of $5$, a person occupying the passage is assigned a cost of $8$, and hard obstructions such as an unmoved chair, cart, or cabinet are assigned a cost of $\infty$. We compare two methods:
\begin{enumerate}
    \item \textbf{Ontology-guided reasoning (ours):} object--affordance candidates are ranked by the utility from Equation~\ref{utility::eq}.
    \item \textbf{Semantic-only baseline:} unresolved objects are ranked by geometric relevance to the nominal corridor while ignoring affordance-state reasoning.
\end{enumerate}

We evaluate the method using Precision@2, Recall@2, feasibility recovery, and the average utility gain of the top-ranked factor. 
Precision@2 measures how many of the top two predicted explanation factors are truly relevant, while Recall@2 measures how many of all relevant factors appear among the top two predictions. 
Feasibility recovery captures whether the identified intervention restores a feasible route, which is particularly important in the library scenario where explanations are intended to help the robot proceed safely. 
Average utility gain measures how strongly the top-ranked explanation factor improves the navigation outcome.

Since there are currently no established state-of-the-art methods for ontology-guided affordance reasoning in navigation explanation, we design the experiments to isolate the core capabilities of the proposed approach, namely identifying relevant object--affordance explanation factors and testing robustness under clutter and incomplete ontology knowledge. 
These experiments should therefore be understood as a first step toward a broader evaluation of this problem.

\section{Results}
We ran the benchmark on 1000 procedurally generated environments with fixed random seeds. 
Table~\ref{tab:main_results} summarizes the main comparison. 
Across all cases, the baseline route was feasible in 45.2\% of environments, while 18.9\% contained at least one shortcut-ground-truth explanation factor. 
On this subset, ontology-guided affordance reasoning achieved a mean Precision@2 of 0.775 and a mean Recall@2 of 0.929, compared with 0.606 and 0.822 for the semantic-only baseline. 
Thus, explicitly reasoning over affordance states improves the identification of route-relevant explanation factors beyond object presence and rough corridor relevance alone.
Furthermore, the ontology-guided method produced, on average, $0.878$ positive explanation candidates per environment. In environments with at least one positive candidate, the top-ranked factor yielded an average utility gain of $597.18$, largely because successful affordance-state changes frequently transformed an initially infeasible route into a feasible one. Together, these results indicate that explicit reasoning over affordance states helps identify explanation factors that are both semantically grounded and practically relevant to navigation continuation.

\begin{table}[h]
\centering
\caption{Results on the procedural robot librarian benchmark.}
\label{tab:main_results}
\begin{tabular}{lcc}
\toprule
Method & Precision@2 & Recall@2 \\
\midrule
Semantic-only baseline & 0.606 & 0.822 \\
Ontology-guided reasoning & \textbf{0.775} & \textbf{0.929} \\
\bottomrule
\end{tabular}
\end{table}

\subsection{Robustness to Semantic Clutter}

We additionally evaluated how both methods behave when the environment contains more semantically irrelevant objects.
Starting from the standard benchmark, we inserted extra distractor objects into free cells of the corridors while keeping the shortcut-relevant explanation factors unchanged. 
We tested whether ontology-guided affordance reasoning remains selective when the scene becomes more cluttered.

We considered five clutter levels by adding 0, 2, 4, 6, and 8 extra distractors per environment. For each level, we generated 1000 procedural environments and again measured Precision@2 and Recall@2 with respect to the shortcut-improving object--affordance pairs. The results are reported in Table~\ref{tab:clutter}.
The results show that ontology-guided reasoning is more robust to semantic clutter than the semantic-only baseline. Without additional distractors, the semantic-only baseline achieved Precision@2 = 0.606 and Recall@2 = 0.822, whereas ontology-guided reasoning achieved Precision@2 = 0.775 and Recall@2 = 0.929. As clutter increased to 8 extra distractors, the baseline degraded to Precision@2 = 0.569 and Recall@2 = 0.763. In contrast, ontology-guided reasoning remained strong, with Precision@2 = 0.877 and Recall@2 = 0.914.

These results suggest that the ontology helps the robot focus on the object--affordance state changes that are genuinely relevant to route recovery. 
This is particularly important in cluttered environments, where many nearby objects may be present, although only a small subset explains why the robot cannot proceed or how the situation could be improved.

\begin{table}[h]
\centering
\caption{Robustness under increasing semantic clutter.}
\label{tab:clutter}
\begin{tabular}{ccccc}
\toprule
\multirow{2}{*}{Extra distractors} & \multicolumn{2}{c}{Semantic-only} & \multicolumn{2}{c}{Ontology-guided} \\
 & P@2 & R@2 & P@2 & R@2 \\
\midrule
0 & 0.606 & 0.822 & 0.775 & 0.929 \\
2 & 0.613 & 0.821 & 0.799 & 0.918 \\
4 & 0.609 & 0.810 & 0.838 & 0.917 \\
6 & 0.592 & 0.790 & 0.833 & 0.910 \\
8 & 0.569 & 0.763 & 0.877 & 0.914 \\
\bottomrule
\end{tabular}
\end{table}

\section{Discussion and Conclusion}
This paper introduced ontology-guided reasoning for affordance-based explanations of robot navigation. 
By representing nearby objects, their affordances, affordance states, and spatial relations in an ontology, the robot can identify explanation factors that are both actionable and relevant to navigation reliability. 
The main advantage of the proposed approach is the identification of semantically meaningful and actionable explanation factors, i.e., object--affordance state changes that would restore or improve the robot's route. 
The ontology provides a structured basis for selecting explanations directly relevant to navigation.

The current benchmark is intentionally lightweight and abstract, and does not yet model perception uncertainty, richer scene graphs, or real-robot execution. 
Future work will extend the framework toward uncertainty-aware ontology updates, learned affordance acquisition, and tighter integration with perception and motion planning. 
An important next step is to deploy and evaluate the framework on a real robot in a library-like environment, under real sensing and execution constraints. 
Another promising direction is to reconnect the present reasoning layer with multimodal visual--textual explanation embodiments.

\newpage
\balance
\bibliographystyle{IEEEtran}
\bibliography{IEEEabrv,references}

\end{document}